\documentclass[letterpaper, 10 pt, conference]{ieeeconf}  

\IEEEoverridecommandlockouts                              

\usepackage{amsmath} 
\usepackage{amssymb}  
\usepackage{dsfont}
\usepackage{graphicx}
\let\labelindent\relax

\usepackage{psfrag,graphicx,epsfig}
\usepackage{epstopdf}
\usepackage{xspace}
\usepackage{subfig}
\usepackage{float}
\usepackage{placeins}
\usepackage{multirow}
\usepackage{pgf,tikz}
\usepackage{nowidow}
\usepackage{lineno}
\usepackage{xcolor}
\usepackage{wrapfig}

\usepackage{siunitx}
\usepackage{color}
\usepackage{flushend}
\usepackage{lineno}
\usepackage{algorithm}
\usepackage[noend]{algpseudocode}
\usepackage{prettyref}
\usepackage{enumitem}
\usepackage{mathtools}
\usepackage{multirow}
\usepackage{bbm}
\usepackage{hyperref}

\usepackage[space]{cite}

\newrefformat{pb}{Problem~\ref{#1}}
\algnewcommand{\LineComment}[1]{\State \(\triangleright\) #1}
\algdef{SE}[DOWHILE]{Do}{doWhile}{\algorithmicdo}[1]{\algorithmicwhile\ #1}

\newcommand{\timeSet}{\mathcal{T}}

\allowdisplaybreaks

\title{\LARGE \bf
Plan-Guided Reinforcement Learning for Whole-Body Manipulation
}

\author{Mengchao Zhang$^{\dagger1}$, Jose Barreiros$^{\dagger2}$, Aykut \"{O}zg\"{u}n \"{O}nol$^{\dagger2}$%
\thanks{$^{\dagger}$ Equal contribution.}  
\thanks{$^1$Department of Mechanical Science and Engineering, University of Illinois at Urbana-Champaign, Urbana, IL, USA 61801. {\tt\small mz17@illinois.edu}}
\thanks{$^2$Toyota Research Institute, Cambridge, MA, USA, 02139. {\tt\small \{jose.barreiros, aykut.onol\}@tri.global}}
\thanks{$^*$This work was done during Mengchao Zhang's internship at Toyota Research Institute.} \thanks{This paper has supplementary downloadable material provided by the authors. It includes an MP4 format movie clip [\href{https://drive.google.com/file/d/1Vu4PwfqgXsMbPwdVNK1pppuQjMOUvIKl/view?usp=sharing}{Link}], which shows demonstrations of our method.}}%

\begin{document}

\maketitle


\begin{abstract}

Synthesizing complex whole-body manipulation behaviors has fundamental challenges due to the rapidly growing combinatorics inherent to contact interaction planning. While model-based methods have shown promising results in solving long-horizon manipulation tasks, they often work under strict assumptions, such as known model parameters, oracular observation of the environment state, and simplified dynamics, resulting in plans that cannot easily transfer to hardware. Learning-based approaches, such as imitation learning (IL) and reinforcement learning (RL), have been shown to be robust when operating over in-distribution states; however, they need heavy human supervision. Specifically, model-free RL requires a tedious reward-shaping process. IL methods, on the other hand, rely on human demonstrations that involve advanced teleoperation methods. In this work, we propose a plan-guided reinforcement learning (PGRL) framework to combine the advantages of model-based planning and reinforcement learning. Our method requires minimal human supervision because it relies on plans generated by model-based planners to guide the exploration in RL. In exchange, RL derives a more robust policy thanks to domain randomization. We test this approach on a whole-body manipulation task on Punyo, an upper-body humanoid robot with compliant, air-filled arm coverings, to pivot and lift a large box. Our preliminary results indicate that the proposed methodology is promising to address challenges that remain difficult for either model- or learning-based strategies alone.


\end{abstract}

\section{Introduction}

\begin{figure}[t]
\centering
\setlength\tabcolsep{1pt}
\includegraphics[width=0.95\linewidth]{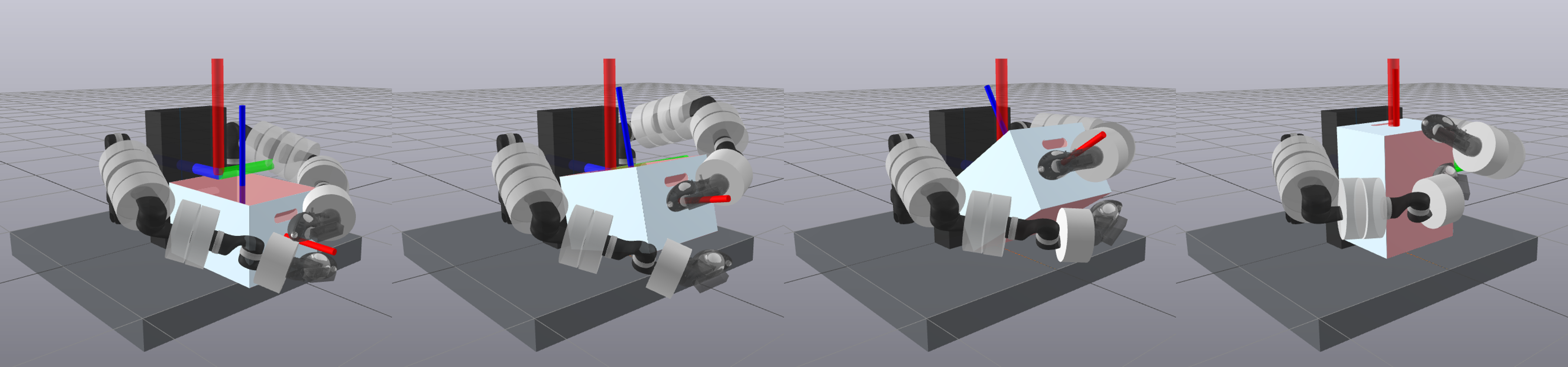}\\
\vspace{-5pt}
\makebox[1.0\linewidth]{\footnotesize (a) GQDP plan}\\
\vspace{5pt}
\includegraphics[width=0.95\linewidth]{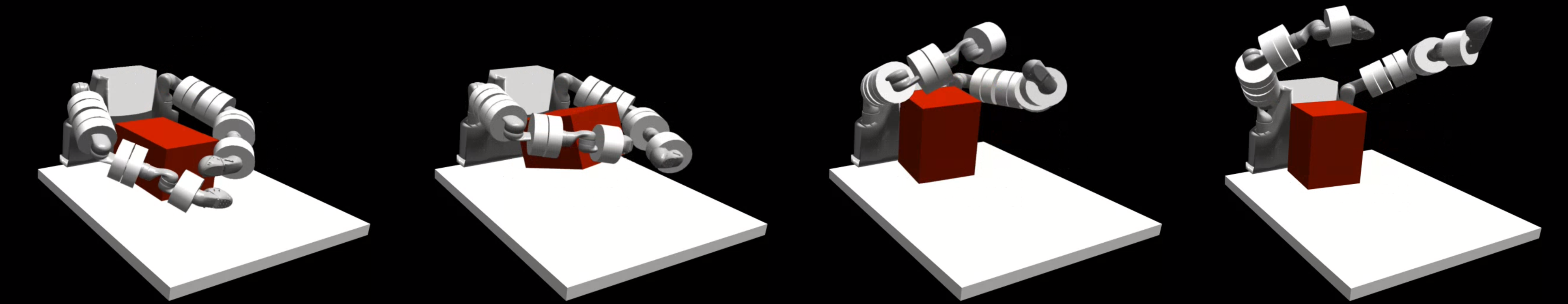}\\
\vspace{-5pt}
\makebox[1.0\linewidth]{\footnotesize (b) RL policy in simulation}\\
\vspace{5pt}
\includegraphics[width=0.95\linewidth]{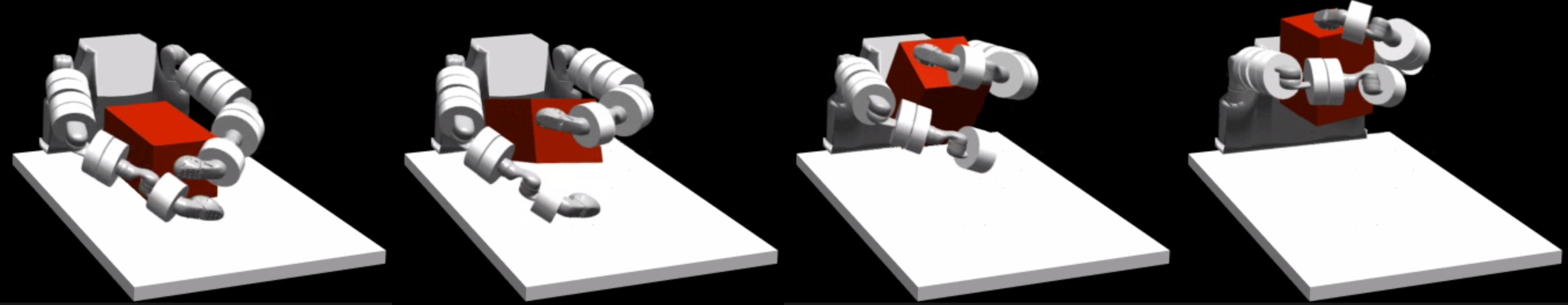}\\
\vspace{-5pt}
\makebox[1.0\linewidth]{\footnotesize (c) PGRL policy in simulation}\\
\vspace{5pt}
\includegraphics[width=0.95\linewidth]{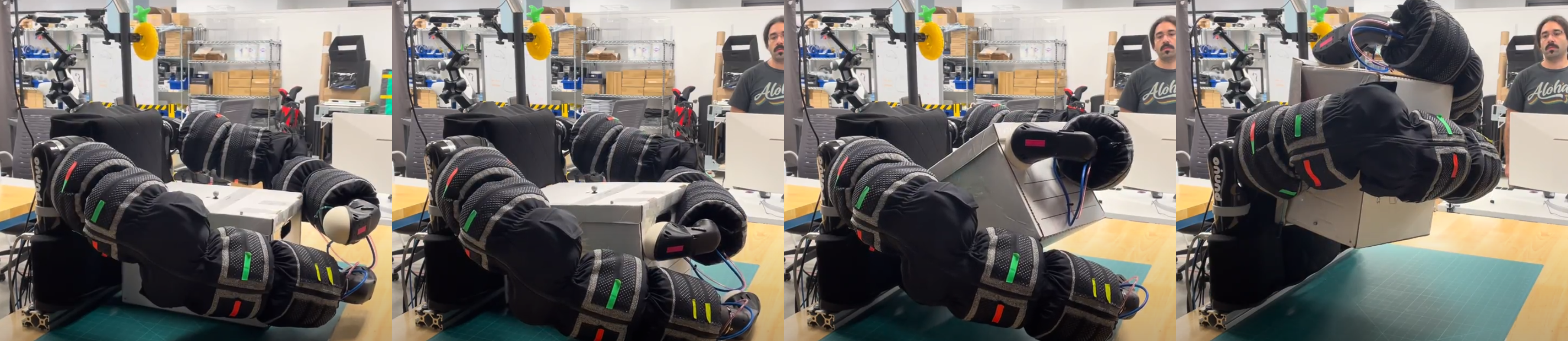}\\
\vspace{-5pt}
\makebox[1.0\linewidth]{\footnotesize (d) PGRL policy on hardware}
\caption{\label{fig:result} Snapshots of the GQDP plan (a) simulation (b, c) and hardware (d) policy rollouts to show the style difference between the plan and the RL and PGRL policies, as well as the effective sim-to-real transfer.}

\end{figure}

Humans employ a diverse array of strategies to effectively manipulate various objects, including dexterous manipulation, full-body engagement, and interactions with the environment \cite{mason2018toward,chavan2020sampling}. In the realm of robotics, there has been a longstanding endeavor to replicate and integrate these intricate human behaviors. For instance, consider the scenario delineated in Fig. \ref{fig:result}(a), wherein a robot is tasked to move a large box to a desired pose. In this pursuit, the robot might initiate the process by employing both arms to pull the box toward its torso, subsequently utilizing a hand to pivot the box by bracing it against its torso until it attains the desired configuration. However, devising a framework to systematically plan and control the execution of such contact-rich behaviors presents a formidable challenge.

Model-based planners struggle with these hybrid-dynamics problems because contacts lead to stiff, non-smooth numerics with an excessive number of discrete contact modes. Although recent advancements in planning with contacts have exhibited promising outcomes in manipulation planning \cite{pang2022global,cheng2022contact,zhang2023simultaneous,natarajan2023torque,onol2020tuning,wang2022contact}, the robust execution of the planned trajectory in hardware remains an unresolved issue. Often, these planners take a considerable amount of time to converge and run offline when trying to discover complex contact sequences. Open-loop execution manifests susceptibility to uncertainties in model parameters and initial pose~\cite{cheng2022contact}, so achieving robust execution necessitates closing the loop.

Imitation learning (IL) emerges as a promising pathway to tackle this challenge, a prospect bolstered by recent progress in gradient-field learning methods~\cite{chi2023diffusion,haldar2023teach,du2023behavior}. Yet, applying this strategy to whole-body manipulation tasks confronts impediments originating from the limitations of teleoperation methodologies. Predominantly tailored for end-effector tracking, these techniques fall short when tasked with effectively showcasing intricate whole-body maneuvers. Even when resorting to whole-body teleoperation techniques (such as motion-capture-based kinematic retargeting) for generating demonstrations, our empirical observations indicate limitations stem from the absence of comprehensive whole-body haptic feedback~\cite{agarwal2022haptics, floreano2018haptics} available to the teleoperator. Furthermore, the substantial volume of demonstrations required to effectively train a proficient imitation learning policy has emerged as another restrictive factor.

Recent advances in reinforcement learning (RL) have yielded remarkable outcomes
in dexterous manipulation~\cite{chen2022system,andrychowicz2020learning,nagabandi2020deep,chen2023visual} that are difficult to replicate with model-based planning. Notably, these advancements frequently hinge upon the availability of task-specific insights, either in the form of well-defined reward functions or expert guidance. The acquisition of such task-related information, however, can pose difficulties, particularly in the domain of whole-body manipulation.

As a means to streamline the process of reward design, guided RL capitalizes on pre-existing knowledge inferred from data to enhance the efficiency and efficacy of the reinforcement learning process~\cite{eesserer2022guided}. In particular, example (or demonstration)-guided RL that aims to combine motion imitation with task-based rewarding (namely, standard RL) has been a popular concept in robotics (as well as in animation) to aid exploration by instilling a desired motion style, accelerate learning, and ease reward shaping~\cite{vecerik2017leveraging,rajeswaran2017learning,peng2018deepmimic,nair2018overcoming,zhu2018reinforcement,zhu2019dexterous,goecks2019integrating,christen2019guided,peng2020learning,nair2020awac,arunachalam2023dexterous}.

Peng et al.~\cite{peng2021amp} recently proposed an example-guided RL framework based on adversarial imitation learning, named Adversarial Motion Priors (AMP). This approach does not require designing imitation objectives or motion selection mechanisms, and it can automatically synthesize a policy that completes a desired high-level task given a set of unstructured example motions. Due to its promise to impose motion characteristics on the RL policy without the burden of reward engineering, the AMP idea which was originally proposed for animation, has caught the attention of the robotics community and led to impressive results for quadruped locomotion, e.g., \cite{vollenweider2023advanced,escontrela2022adversarial,li2023learning,li2023versatile,wu2023learning}. These are compelling examples showing that the AMP approach can generate policies: (i) with a desired style even from infeasible, partial, and scarce demonstrations; and (ii) that can directly transfer to the real world.

However, producing demonstrations through teleoperation for intricate whole-body manipulation tasks remains a challenging endeavor due to relying on human supervision, which hinders the potential of this method to scale to the large set of diverse skills needed for a robot to operate in the real world. As a potential solution, state-of-the-art global-planning-through-contact methods, such as the Global Quasi-Dynamic Planner (GQDP)~\cite{pang2022global}, can synthesize long-horizon behaviors with complex intermittent contact interactions, yet the resulting plans are fragile (even with a perfect model) when executed in an open-loop fashion. Our hypothesis is that AMP can help us convert a rough (and potentially infeasible) plan into a feedback policy that can immediately be deployed on hardware. It is noteworthy that while it is possible to track plans with conventional control methods that often entail estimations for hidden model parameters, learning approaches let us work directly with output feedback.

Imitating plans is not a new idea and has been previously studied, especially for quadruped locomotion, e.g., \cite{bogdanovic2022model,fuchioka2023opt,miller2023reinforcement,kang2023rl}. In the context of manipulation, Wang et al.~\cite{wang2022goal} proposed using a motion and grasp planner for two purposes: (i) to generate demonstrations for off-policy learning and (ii) to supervise an auxiliary-goal, grasp-pose-prediction task. As a result, they obtain closed-loop grasping and human-robot handover behaviors. Huang et al.~\cite{huang2023diffusion} propose a diffusion-based generative sampling framework that can provide physics-aware goal-oriented plans given a 3D scene. They showcase this approach for dexterous grasp generation and collision-aware arm motion planning using offline-generated plans. However, due to its pure imitation nature, this method requires several high-fidelity plans.

\begin{figure}[t]
\centering
\setlength\tabcolsep{1pt}
\renewcommand{\arraystretch}{0.0}
\includegraphics[width=0.9\linewidth]{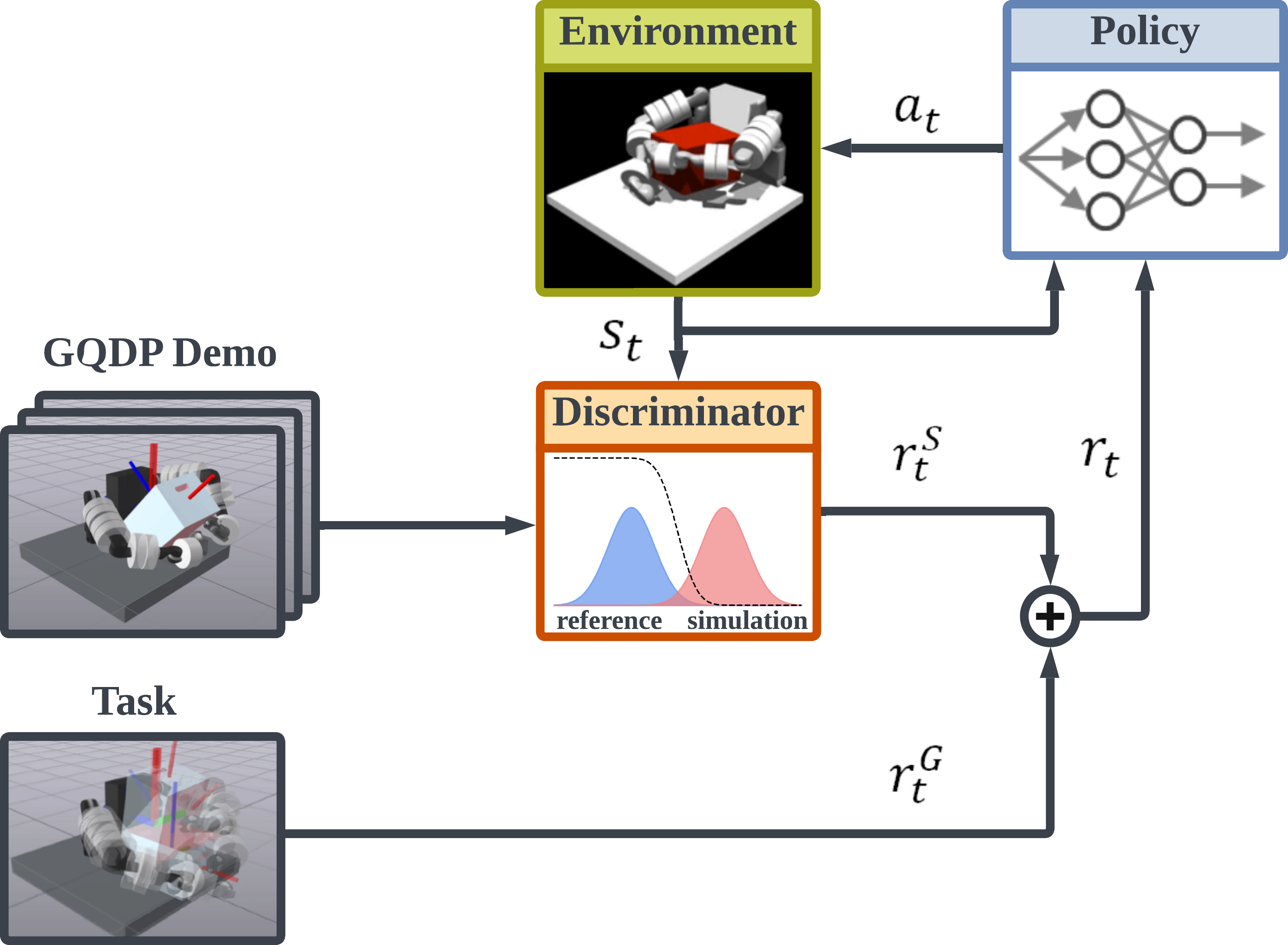}
\caption{\label{fig:flowchart} A flowchart of the proposed framework: the process commences with the acquisition of a reference motion dataset, generated through the utilization of a model-based planner; then the system undertakes the training of a discriminator, designed to acquire proficiency in learning an imitation reward, and a policy, which in turn empowers the robot to replicate the demonstrated motion style while simultaneously accomplishing the intended task.}
\vspace{-10pt}
\end{figure}

Despite exciting advancements in leveraging planning to guide RL for locomotion problems, this has not yet translated to dexterous manipulation, probably due to the unavailability of reliable planners until recently. In this work, we introduce a simple but effective framework named Plan-Guided Reinforcement Learning (PGRL), which is illustrated in Fig. \ref{fig:flowchart}. This framework, while planner agnostic, first uses the GQDP to synthesize quasi-dynamic plans which are not necessarily feasible, given the desired pose of the manipuland. Then, the AMP is used to complete the gaps in the plan and derive a closed-loop policy with similar motion characteristics to the plan. We test this approach to address a complex whole-body manipulation task employing Toyota Research Institute's Punyo robot~\cite{goncalves2022punyo}. Our preliminary findings show that this approach can enable generating policies efficiently even from a single, infeasible example plan. Notably, our method obviates the requirement for human demonstrations and operates in conjunction with a simplified reward function. Facilitated by domain randomization and the inherent robustness stemming from Punyo's passive compliance, the trained policies can be seamlessly transferred to real hardware without necessitating subsequent processing. While we present an instance of this approach using GQDP, we emphasize that this approach is not limited to a certain source for the demonstration and could be used with other types of planners and teleoperation data. Nevertheless, the impact of the demonstration type on the performance is an open question.

\section{Method}
In this section, we briefly describe the methods we use for the proposed framework. The implementation details can be found in section \ref{sec:training_details}.

\subsection{Planning Through Contact}
\label{sec:gqdp}
We select GQDP~\cite{pang2022global} as the contact-implicit planner because of its capability for synthesizing long-horizon behaviors with multiple intermittent contacts. This approach assumes quasi-dynamics to reduce the problem into the configuration space and uses a contact smoothing scheme that along with the locally linear model is used to derive a reachability metric. As a result, the planning through contact problem given initial and desired configurations can be solved by a sampling-based planner, in this case, the rapidly exploring random tree (RRT) method \cite{lavalle1998rapidly}. The path generated by RRT is then refined using trajectory optimization to output a trajectory denoted as $\timeSet'=\{(\mathbf{q}^{a}_{t}, \mathbf{q}^{u}_{t}, \mathbf{a}_{t})|{t=0}, \cdots, T\}$. Here, $\mathbf{q}^{a}_{t}$ and $\mathbf{q}^{u}_{t}$ are the configurations of the actuated and unactuated degrees of freedom of the system, and $\mathbf{a}_{t}$ denotes the robot position commands (or actions) for a joint stiffness controller at each discrete time step $t$.

It is pertinent to note that the resulting plan may exhibit non-physical behavior because of: (i) the quasi-dynamic assumption breaking when the system moves at non-negligible speeds and (ii) a robot teleport issue when switching contact modes. Due to (ii), the manipuland is not guaranteed to remain stable when the robot transitions from one contact configuration to another. This entails that while the robot follows the planned waypoints $\mathbf{q}^{a}_{t}$, the manipuland may not stay at the corresponding $\mathbf{q}^{u}_{t}$ for unstable tasks. Consequently, in our approach, we retain only the robot's configuration sequence $\timeSet=\{\mathbf{q}^{a}_{t}|{t=0}, \cdots, T\}$ as the demonstration.

\subsection{Example-Guided Reinforcement Learning}
Given a dataset of reference motions and a task objective defined by a reward function, AMP synthesizes a control policy that enables the agent to achieve the task objective, while utilizing behaviors that resemble the motion style of the dataset. It is important to emphasize that the agent's behaviors are not obligated to precisely replicate individual motions from the dataset. Rather, the agent is encouraged to embody the broader characteristics observed within the collection of reference motions. This renders the GQDP a promising candidate for serving as the demonstrator. The emphasis here lies not on the successful task completion through the plan, but rather on utilizing it as a stylistic guide to steer the robot's behavior and increase the training efficiency and effectiveness.

Our policy is trained through an RL framework, wherein an agent engages with an environment in accordance with a policy denoted as $\pi$. At each time step $t$, the agent perceives the state $\mathbf{s}_t \in\mathcal{S}$ of the system. It then proceeds to sample an action $\mathbf{a}_t\in\mathcal{A}$ from the policy, adhering to the probabilistic distribution $\mathbf{a}_t \sim \pi(\mathbf{a}_t|\mathbf{s}_t)$. Subsequently, the agent runs this chosen action, leading to a new state $\mathbf{s}_{t+1}$, accompanied by a scalar reward $r_t=r(\mathbf{s}_t,\mathbf{a}_t,\mathbf{s}_{t+1})$. In accordance with \cite{peng2021amp}, we formulate the reward function as comprising two distinct components: (i) the task reward $r^G$ that quantifies the degree of task accomplishment, and (ii) the style reward $r^S$ that assesses the resemblance between the robot's motion and the reference motions:
\begin{equation}
r(\mathbf{s}_t,\mathbf{a}_t,\mathbf{s}_{t+1}) = \lambda r^G(\mathbf{s}_{t+1},\mathbf{a}_t) + (1 - \lambda) r^S(\mathbf{s}_t,\mathbf{s}_{t+1}),
\nonumber
\end{equation}
where $\lambda \in [0, 1]$ determines the task reward weight with respect to the imitation reward weight.

To minimize reward shaping, we choose to employ a straightforward task-reward function $ r^G = d_{trans} + d_{rot} + p$, where $d_{trans} $ and $d_{rot}$ quantify the translation and rotation distances to the goal manipuland pose, and $p$ encompasses conventional penalty components often integrated for RL, such as a penalty for actions and velocities.

$r^S$ is decided by the output of a discriminator, which discriminates whether a state transition belongs to the demonstration distribution or not, and the discriminator is trained together with the policy from scratch. Prior to presenting a state transition as input to the discriminator, an observation map $\Phi(\mathbf{s}_t)$ extracts the features from the system state $\mathbf{s}_t$ for selecting the attributes associated with a particular skill. This is an important aspect since it lets us get away with plans that contain infeasible actions by only mimicking the equilibrium joint poses $\mathbf{q}_a$ for imitation, unlike typical behavior cloning methods that imitate feasible actions.

\section{Experiments \& Results}

\begin{figure}[b]
\centering
\setlength\tabcolsep{1pt}
\renewcommand{\arraystretch}{0.0}
\includegraphics[width=0.8\linewidth]{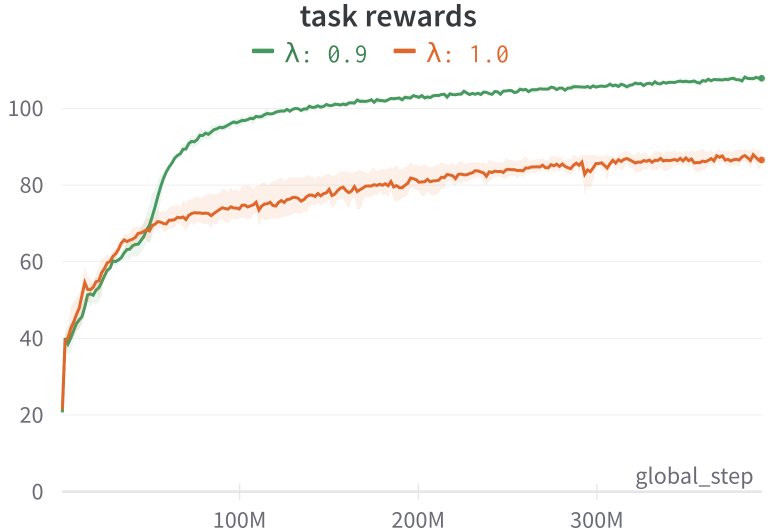}
\caption{\label{fig:reward} Performance of PGRL ($\lambda=0.9$) and RL ($\lambda=1$) where $\lambda$ is the task reward weight.}
\end{figure}

\begin{figure*}[t]
\centering
\setlength\tabcolsep{1pt}
\renewcommand{\arraystretch}{0.0}
\includegraphics[width=0.7\linewidth]{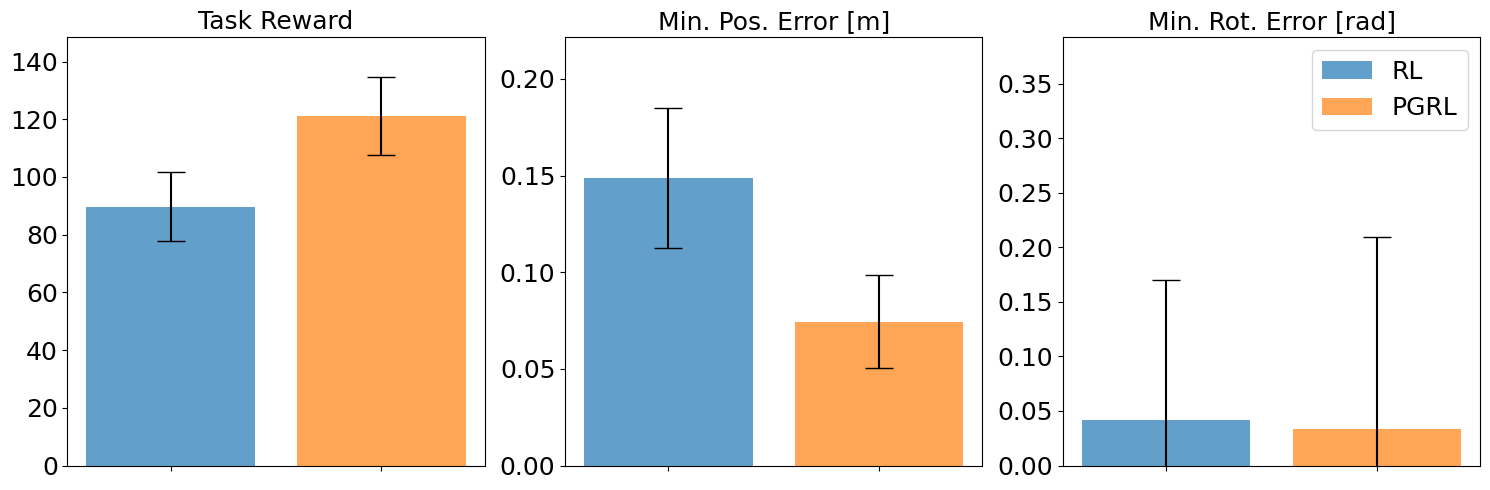}
\caption{\label{fig:results} Aggregated results of rolling out the RL and PGRL policies in 1000 different environments. The box plots show the mean and standard deviation for the task reward and the minimum translational and rotational distances of the manipuland to the goal along each rollout.}
\vspace{-20pt}
\end{figure*}

We test our approach on a whole-body manipulation task for pivoting and lifting a box, as depicted in Fig. \ref{fig:result}(a). This evaluation is conducted using the Punyo robot~\cite{goncalves2022punyo}: an assemblage comprising of two Jaco robot arms, each with soft visuotactile sensors as end-effectors~\cite{kuppuswamy2020soft} and an array of 7 air-filled pressure-based sensors that cover each arm. To estimate the manipuland pose, we use an OptiTrack motion-capture system.

The code provided in \cite{pang2022global} is used for planning. Then, the refined plan obtained from this framework is post-processed for two reasons: (i) to prevent undesired robot-manipuland collisions and (ii) to stabilize the object using a task-specific heuristic that makes the idle left arm to support the box when moving between segments. For this purpose, the kinematic trajectory optimization method in Drake~\cite{drake} is used with minimum distance constraints.

For policy learning, we employ the proximal policy optimization (PPO) algorithm~\cite{schulman2017proximal}. Particularly, we utilize the PPO implementation in \cite{rl-games2021} and the AMP implementation in \cite{IsaacGymEnvs} to run in Isaac Gym \cite{makoviychuk2021isaac}. Furthermore, we incorporate domain randomization \cite{tobin2017domain} to facilitate the transfer of learned behaviors from simulation to real. See the Appendix section for further implementation details.

We run simulation experiments to compare PGRL to GQDP and to standard model-free RL (denoted as RL), which is also trained with PPO using only the task reward. Additionally, we test the policy generated by our proposed approach (PGRL) on hardware. When rolling out the GQDP plan in simulation, the robot encounters difficulty in successfully lifting the box due to the teleportation issue described in section \ref{sec:gqdp}. The inability to achieve success in this scenario in the absence of any model mismatch from planning to simulation prompted the decision not to proceed with testing under conditions of likely model-mismatch in real.

Figure \ref{fig:result} presents a juxtaposition between key points extracted from the plan employed for training and analogous key points identified during the execution of the RL policies. See the accompanying video [\href{https://drive.google.com/file/d/1Vu4PwfqgXsMbPwdVNK1pppuQjMOUvIKl/view?usp=sharing}{Link}] for the full motions. The visual analysis shows the proficiency of our method in emulating the motion style imposed by the plan, whereas standard RL leads to unnatural behaviors with unsatisfactory task performance. Specifically, the RL policy can only pivot the box but fails to maintain it at the desired height. Moreover, the RL training in PGRL helps to bridge the gaps in the GQDP plan (i.e., infeasible transitions during the teleports) and enables task completion. It is also noteworthy that PGRL alters the contact sequence in the plan to a more robust one owing to domain randomization.

To compare PGRL and RL, we train policies by maintaining uniformity across all parameters except for $\lambda$, which is set to 1 (i.e., no imitation reward) for the RL case. Fig. \ref{fig:reward} illustrates the mean task reward curve along with the standard deviation resulting from three training instances with different seeds. Notably, our approach (trained with $\lambda=0.9$) outperforms RL in terms of task reward even though RL's sole objective is to maximize this metric.

To evaluate the trained policies, we roll them out in 1000 distinct environment instantiations, encompassing variations in the physical properties of the manipuland (such as mass, friction, and size) as well as the initial position (extracted from the identical distribution utilized during training). The mean and standard deviation of the accumulated task reward, and the minimum translational and rotational distances of the manipuland to its goal pose along each rollout for both RL and PGRL are shown in Fig. \ref{fig:results}.




\section{Conclusion \& Future Work}

In this study, we introduced a plan-guided RL framework to convert complex, long-horizon, contact-rich manipulation trajectories to a closed-loop policy that is sufficiently robust to be deployed in the real world. As a proof of concept, we tested the proposed approach on a whole-body box reorientation and lifting task using a single, infeasible plan. The results imply that: (i) PGRL can fix infeasible transitions in the plan while maintaining the desired motion characteristics, (ii) the guidance can help to compose a strategy that is better than one that standard RL can discover alone without extensive reward engineering, and (iii) the RL training and domain randomization can enable obtaining a similar but more robust behavior compared to the plan by altering the contact configurations as necessary.

The preliminary results are promising but there are many open research questions that require further investigation. In particular, our future studies would explore the following:
\begin{itemize}
    \item incorporating tactile information into the imitation and/or policy observation spaces with the hope that this can mitigate the need for motion capture for tracking the manipuland's pose (which is already restrictive due to heavy occlusions during whole-body interactions) and enable extending the framework to tasks that are more sensitive to force interactions;
    \item the impact of the quality and quantity of example motions (either plans or teleoperation data with different fidelity) on the performance;
    \item synthesizing more complex behaviors by using examples for diverse skills with the hope that AMP would be able to automatically compose them;
    \item understanding the effect of the demonstrated contact sequence on the exploration and how it is modified during the RL training; and
    \item trying methods like curriculum learning to increase learning efficiency and policy robustness.
\end{itemize}

\newpage
\clearpage

\bibliographystyle{IEEEtran}
\bibliography{references}

\section{Appendix}

\subsection{Implementation Details}
\label{sec:training_details}
\subsubsection{Task Reward}
The task reward is implemented as 
\begin{align}
    \begin{split}
    r^G_t = w_{trans} (1/(\|d_{trans}(\mathbf{q}^u_t,\mathbf{q}^u_{goal})\|+0.1)) + \\    w_{rot} (1/(\|d_{rot}(\mathbf{q}^u_t,\mathbf{q}^u_{goal})\|+0.1)) + \\
    w_{action} ||\mathbf{a}_t||^2 + w_{velocity} ||\dot{\mathbf{q}}^u_t||^2 + w_{termination} \mathbbm{1}(\mathbf{q}^u_t).
    \end{split}
    \nonumber
\end{align}
The initial two terms incentivize task completion. The following two terms impose penalties on both the robot's actions and the manipuland's velocity. The final term enforces a penalty upon the activation of termination conditions. These conditions manifest when the box deviates significantly from the center of the table or experiences a drop. The function $\mathbbm{1}(\cdot)$ is an activation function based on termination conditions. We use the weights $w_{trans}=0.07$, $w_{rot}=0.03$, $w_{action}=-0.002$, $w_{velocity}=-0.002$, and $w_{termination}=-1$ for all experiments. The task is defined by $\mathbf{q}_{goal}^u=(0.15,0,0.4,0,-\pi/2,0)$ concatenating the positions in m and roll-pitch-yaw in rad.

\subsubsection{Observations}
The observation of the discriminator is joint position transitions of the robot ($\mathbf{q}^a_{t}$, $\mathbf{q}^a_{t+1}$). The observation of the policy is ($\mathbf{q}^a_t,\mathbf{q}^u_t,\mathbf{p}^{ee}_t$) where $\mathbf{p}^{ee}_t$ is the Cartesian end-effector poses and $\mathbf{q}^u_t$ is the manipuland pose.

\subsubsection{Training Parameters}

The learning networks and algorithm were implemented in PyTorch 1.8.1 with CUDA 12.0. The training procedure encompassed the collection of experiences from 4096 uncorrelated instances of the simulator performed in parallel. The entirety of the experimental work was executed on a desktop equipped with NVIDIA 3090 GPUs. A single run comprising 1,500 iterations, adhering to the aforementioned computational settings and device specifications, was accomplished within approximately 4 hours. Detailed training parameters and network architectures are outlined in Tables \ref{tab:training_parameters} and Table \ref{tab:nn_parameters}, respectively.

\begin{table}[H]
\centering
\begin{tabular}{cc}
\hline
Parameter                      & Value                      \\ \hline
$\gamma$                       & 0.99                       \\
$\tau$                         & 0.95                       \\
$\lambda$                      & 0.9                        \\
parallel training environments & 4096                       \\
sample per update iteration    & 4096                       \\
batch size                     & 512                        \\
learning rate                  & $5e^{-5}$ \\
KL divergence target           & $8e^{-3}$ \\
clip range                     & 0.2                        \\
horizon length                 & 64                         \\
discriminator weight decay     & $1e^{-4}$ \\
discriminator gradient penalty & 10                         \\ \hline
\end{tabular}
\caption{Training parameters.\label{tab:training_parameters}}
\end{table}

\begin{table}[H]
\centering
\begin{tabular}{cccc}
\hline
Network        & Type & Hidden Layers            & Activation                   \\ \hline
policy         & MLP  & [256, 128, 64] & Rectified Linear Unit (ReLU) \\
value function & MLP  & [256, 128, 64]  & Rectified Linear Unit (ReLU) \\
discriminator  & MLP  & [256, 128, 64]  & Rectified Linear Unit (ReLU) \\ \hline
\end{tabular}
\caption{Network architecture.\label{tab:nn_parameters}}
\end{table}

\subsubsection{Domain Randomization}

Domain randomization techniques are judiciously employed during the training process to enhance robustness and performance. This involves the systematic application of disturbances to various parameters. Specifically, we initialize the manipuland on the table uniformly within a 5 cm radius of its nominal initial position [0.35. 0, 0.13] m. Additionally, a perturbation sampled from the uniform distribution $\mathcal{U}(-0.05,0.05)$ is added to its yaw angle. A noise sampled from $\mathcal{N}(0,0.02)$ rad is injected into the robot's actions (i.e., joint position commands to a stiffness controller), and a disturbance drawn from $\mathcal{U}(0.0,0.5)$ m/s$^2$ is added to the gravity. Furthermore, the distributions $\mathcal{U}(0.9,1.1)$, $\mathcal{U}(0.8,1.2)$, and $\mathcal{U}(0.8,1.0)$ are used to scale the dimensions and mass of the manipuland and the friction coefficient for all contacts. The nominal manipuland dimensions and mass are [0.41. 0.315, 0.26] m and 0.45 kg, and the nominal friction coefficient is 1.

\subsection{Acknowledgments}
The authors extend their sincere appreciation to Tao Chen, Chenhao Li, and Binghao Huang for their valuable assistance in the implementation process. Gratitude is also expressed to Russ Tedrake, Hongkai Dai, Benjamin Burchfiel, and the Dexterous Manipulation group at TRI for their insightful discussions. Furthermore, the authors would like to thank Alex Alspach, Sam Creasey, Aimee Goncalves, and the entire Punyo team for their generous support, constructive feedback, and the exceptional hardware provided.

\end{document}